\documentclass[12pt,letterpaper]{article}
\usepackage[a4paper, total={7in, 10in}]{geometry}

\usepackage{graphicx}
\usepackage{helvet}
\usepackage{authblk}
\usepackage{hyperref}
\usepackage{amsmath} 
\usepackage{amssymb} 
\usepackage{orcidlink} 
\usepackage[super,comma,sort&compress]  
   {natbib}
\usepackage[right]{lineno} \linenumbers
\usepackage[title]{appendix}%

\usepackage[utf8]{inputenc} 
\usepackage[T1]{fontenc}    
\usepackage{url}            
\usepackage{booktabs}       
\usepackage{amsfonts}       
\usepackage{nicefrac}       
\usepackage{microtype}      
\usepackage{xcolor}         
\usepackage{multirow}
\usepackage{threeparttable}
\usepackage{amsmath}
\usepackage{subcaption}
\usepackage{makecell}
\usepackage{placeins}
\usepackage[utf8]{inputenc} 
\usepackage{booktabs}      
\usepackage{siunitx}       
\usepackage{caption}       
\usepackage{tabularx}
\usepackage{listings}    
\usepackage{pifont}
\usepackage{array}
\usepackage{makecell}
\usepackage[markup=underlined]{changes} 
\usepackage{threeparttable}  
\usepackage{bm}

\renewcommand{\color}[1]{} 

\theadset{\bfseries}

\makeatletter
\renewcommand{\maketitle}{\bgroup\setlength{\parindent}{0pt}
\begin{flushleft}
  \textbf{\@title}
  
  \@author
\end{flushleft}\egroup}
\makeatother

\title{Masked Training for Robust Arrhythmia Detection from Digitalized Multiple Layout ECG Images}
\date{}

\author[1,7,\orcidlink{0009-0000-1986-8131}]{Shanwei Zhang}
\author[2,\orcidlink{0000-0002-4041-3083}]{Deyun Zhang}
\author[3]{Yirao Tao}
\author[1]{Kexin Wang}
\author[2]{Shijia Geng}
\author[7,8]{Jun Li}
\author[4]{Qinghao Zhao}
\author[3]{Xingpeng Liu}
\author[6]{Xingliang Wu}
\author[1]{Shengyong Chen}
\author[1,5,*]{Yuxi Zhou}
\author[7,8,9,10,*,\orcidlink{0000-0001-7521-5127}]{Shenda Hong}

\affil[1]{Department of Computer Science, Tianjin University of Technology, Tianjin, China}
\affil[2]{HeartVoice Medical Technology, Hefei, China}
\affil[3]{Department of Cardiology, Beijing Chaoyang Hospital, Capital Medical University, Beijing, China}
\affil[4]{Department of Cardiology, Peking University People’s Hospital, Beijing, China}
\affil[5]{DCST, BNRist, RIIT, Institute of Internet Industry, Tsinghua University, Beijing, China}
\affil[6]{Tianjin Key Laboratory of Ionic-Molecular Function of Cardiovascular Disease, Department of Cardiology, Tianjin Institute of Cardiology, the Second Hospital of Tianjin Medical University, Tianjin 300211, China}
\affil[7]{National Institute of Health Data Science, Peking University, Beijing, China}
\affil[8]{Institute for Artificial Intelligence, Peking University, Beijing, China}
\affil[9]{Institute of Medical Technology, Peking University Health Science Center, Beijing, China}
\affil[10]{State Key Laboratory of Vascular Homeostasis and Remodeling, NHC Key Laboratory of Cardiovascular Molecular Biology and Regulatory Peptides, Peking University, Beijing, China}

\affil[*]{Correspondence: joy\_yuxi@pku.edu.cn,hongshenda@pku.edu.cn}

\begin{document}

\maketitle

\section*{ABSTRACT}

{\color{blue}
\subsection*{Background}
Electrocardiograms are indispensable for diagnosing cardiovascular diseases, yet in many clinical and resource-limited settings they exist only as paper printouts or photographs stored in multiple recording layouts. Converting these images into digital signals introduces two key challenges: temporal asynchrony among leads and partial blackout missing, where contiguous signal segments become entirely unavailable. Existing models cannot adequately handle these concurrent problems while maintaining interpretability.
\subsection*{Methods}
We propose PatchECG, combining an adaptive variable block count missing learning mechanism with a masked training strategy. The model segments each lead into fixed-length patches, discards entirely missing patches, and encodes the remainder via a pluggable patch encoder. A disordered patch attention mechanism with patch-level temporal and lead embeddings captures cross-lead and temporal dependencies without interpolation. PatchECG was trained on the PTB-XL dataset and evaluated under seven simulated layout conditions, with external validation on 400 real ECG images from Chaoyang Hospital across three clinical layouts 
(12$\times$1, 3$\times$4, and 6$\times$2).
\subsection*{Results}
PatchECG achieves a consistent average area under the receiver operating characteristic curve of approximately 0.835 across all simulated layouts. On the Chaoyang Hospital cohort, the model attains an overall area under the receiver operating characteristic curve of 0.778 for atrial fibrillation detection, rising to 0.893 on the 12$\times$1 subset—surpassing the large-scale pre-trained baseline by 0.111 and 0.190, respectively. Model attention aligns with cardiologist annotations at a rate approaching inter-clinician agreement.
\subsection*{Conclusions}
PatchECG provides a robust, interpolation-free, and interpretable solution for arrhythmia detection from digitized ECG images across diverse layouts. Its direct modeling of asynchronous and partially missing signals, combined with clinically aligned patch-level attention, positions it as a practical tool for cardiac diagnostics from legacy ECG image archives in real-world clinical environments.
}


\section*{Plain language summary}
{\color{blue}
ECG recordings of the heart's electrical activity are a cornerstone of cardiac diagnosis. In many hospitals worldwide, ECGs exist only as paper printouts or photographs taken in different layouts. Converting these images into digital signals introduces missing and misaligned data that existing models cannot reliably handle. We developed PatchECG, a model that works directly with incomplete and misaligned ECG signals across any recording format, without artificially filling in the gaps. Instead of patching over missing data, PatchECG learns to focus on the signal segments that matter most—the same features experienced cardiologists examine. Validated on real hospital images, PatchECG consistently outperformed existing methods and produced interpretable, clinically meaningful outputs. This work offers a practical pathway to unlock decades of paper ECG archives for automated cardiac diagnosis, with particular relevance for under-resourced healthcare settings.
}

\section*{Introduction}

Cardiovascular disease (CVD) is one of the world's deadliest disease, causing approximately 17.8 million deaths according to 2017 statistics \cite{mensah2019global}. Electrocardiogram (ECG) is a key tool for the diagnosis and treatment of CVD, revealing important pathophysiological information by recording the electrical activity of the heart. In recent years, deep learning techniques have been shown to rival trained doctors in ECG signal and ECG arrhythmia detection tasks. Despite recent advances in digital ECG equipment, the use of photographs, printed screens, or paper ECGs is still prevalent in the Global South, where ECG slices are far less expensive to store than signal data, and where these recordings not only contain a rich history of CVD, but also reflect its diversity. In underdeveloped regions, this non-digitized data format highlights the urgent need for high-quality digitization and automated analysis tools \cite{Goldberger2000}. Thus, digitizing traditional ECG images and building algorithms capable of automated diagnosis based on them will both help to deepen our understanding of CVD and, hopefully, significantly improve the diagnosis and treatment of underrepresented and medically under-resourced populations. 

{\color{blue}
In clinical practice, when AI models are applied to ECG image data, image-based classification methods suffer from a fundamental interpretability limitation: their predictions are often driven by background features rather than ECG waveform morphology, which undermines clinician trust. In contrast, signal-based models can provide waveform-level explanations that directly correspond to established diagnostic criteria (e.g., P-wave morphology, RR interval regularity). By digitizing ECG images into signals and applying PatchECG, clinicians gain access to interpretable, signal-level diagnostic support from legacy image archives—particularly valuable in settings where only paper or photographic ECG records exist.}

\begin{figure*}[]
\centerline{\includegraphics[width=1\textwidth]{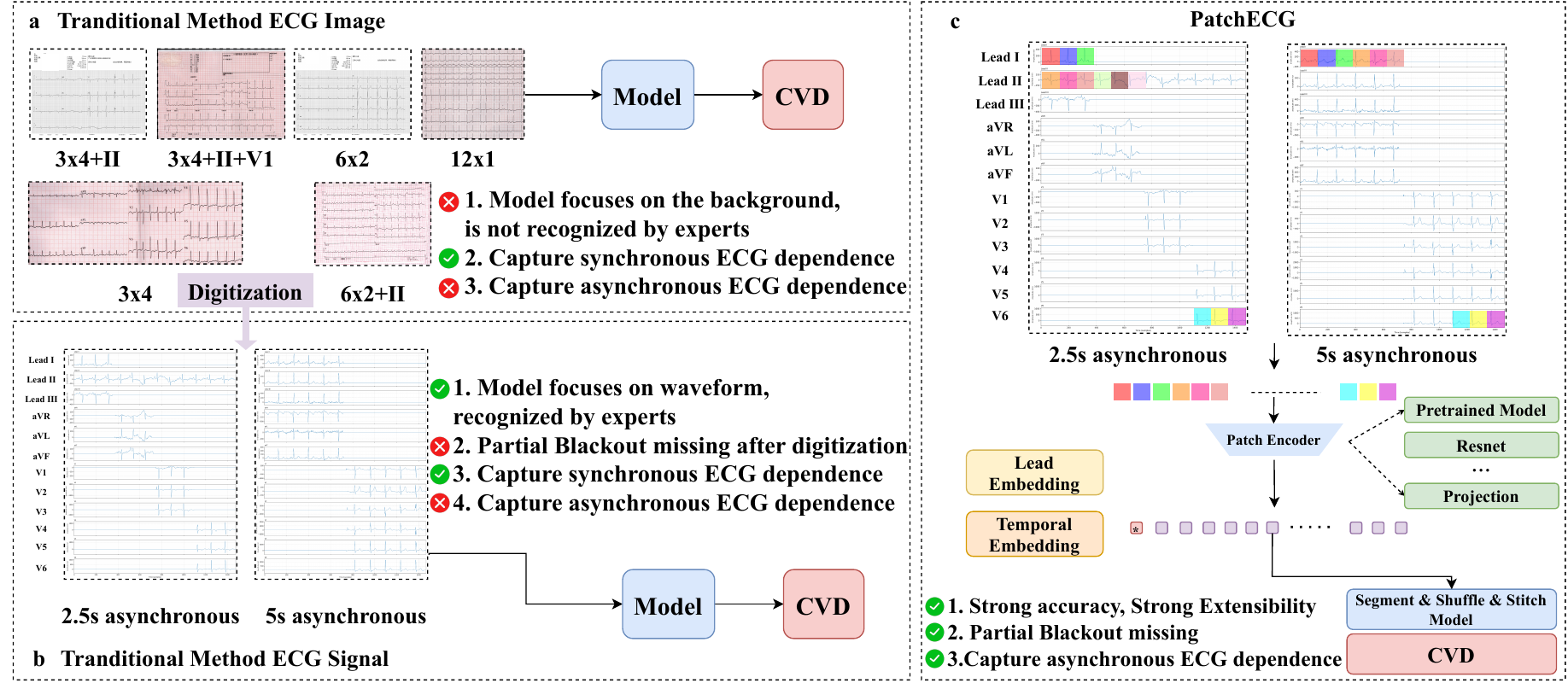}}
\caption{ \textbf{a}: Traditional image methods focus on the image background and are limited in accuracy, and cannot effectively capture asynchronous ECG dependencies. \textbf{b}: Signal methods are highly accurate, cannot capture asynchronous asynchronous ECG dependencies and effectively deal with partial blackout missing. \textbf{c}: We propose a variable block number mechanism and patch guided attention mechanism based on masking training adaptation, which can effectively handle ECG with different layouts.}
\end{figure*}
ECG images exist in a variety of layouts. For example, in Fig. 1(a), different forms such as common 3 x 4 + \uppercase\expandafter{\romannumeral 2} layout, 6 x 2 layout or complete 12-lead may be presented. In existing studies, ECG images with different layouts are usually directly inputted into an image classification model for classification\cite{sangha2022automated,gliner2025clinically,yang2025ecg}. Data augmentation (e.g., rotation, folding, light and shadow adjustments, disrupting lead positions, etc.) of such images can effectively improve the robustness of the model so that it still maintains a high accuracy in predicting real-world data.  However, image classification models are prone to focus excessively on the image background rather than the ECG signal itself. Although ECG all-in-one \cite{gliner2025clinically} provides more accurate interpretability by calculating the classifier gradient through Jacobi matrix sensitivity, its interpretation results still focus on the image background. This tends to raise questions from experts about its interpretability, as physicians have more confidence in the prediction results based on the ECG signal itself than the image model.

With the development of methods for digitizing ECG images, ECG image data can be accurately converted into ECG signal data\cite{krones2024combining,fortune2022digitizing}, making it possible to use the signal data for subsequent analysis. As shown in Fig. 1(b), the digitization of ECG image with different layouts of image produces 2.5s asynchronous, 5s asynchronous and synchronous signals \cite{sau2025comparison} and “Partial Blackout” which covers situations where a variable number of features become unavailable for some time\cite{islam2025self}. This digitized signal ECG presents two major challenges:

\begin{itemize}
\item There are differences in the way ECG images are stored in different hospitals, such as the common 3×4 or 6×2 layouts. After digitizing these images, the signal data are often missing to varying degrees of partial blackout. It remains a challenging task to effectively model multi-lead ECG data with partial blackout without relying on additional interpolation (avoid introducing noise) and preserving the original information as much as possible.
\item Existing methods usually assume that multilead ECG signals are synchronized; however, after digitization of different layouts of ECG images, the signals of each lead appear to be inconsistently missing in time, resulting in asynchronous ECGs of 2.5s or 5s. Current methods, whether based on image or signal level, are difficult to capture the synergistic dependencies and temporal features among multiple leads in this asynchronous mode, and achieving effective modeling is still a challenge to be solved.
\end{itemize}

To overcome these challenges, We propose PatchECG—a framework based on a masked training strategy and not relying on additional interpolations that can be adapted to different layouts of ECGs. The main contributions of this study are as follows:

\begin{itemize}
    \item Aiming at the missing problem due to different layouts, we design an adaptive variable block number missing learning mechanism. Combined with a masked training strategy, it is able to directly model the missing regions without relying on any interpolation operation, effectively capturing the temporal features and synergistic dependencies among multiple leads.
    \item In order to capture the complex structure of ECG with different layouts, we construct a disordered patch attention mechanism to encode patches and introduce patch-level temporal-lead embedding. We automatically focus on the important segments that have synergistic dependencies between leads in terms of timing, thus realizing the intelligent recognition of arrhythmia in any ECG layout.
    {\color{blue}
    \item We validate PatchECG on a real-world clinical cohort of 400 ECG images from Chaoyang Hospital, covering three common clinical layouts (3$\times$4, 6$\times$2, and 12$\times$1). PatchECG achieved an overall AUROC of 0.778 for atrial fibrillation diagnosis across all layouts, and 0.893 on the 12$\times$1 subset, outperforming the large-scale pre-trained model ECGFounder by 0.111 and 0.190 respectively. Furthermore, quantitative interpretability evaluation against two experienced cardiologists confirms that the model's attention aligns with clinically meaningful ECG segments.}

\end{itemize}

\section*{Methods}


\subsection*{Related Works}
\subsubsection*{Perform Prediction after Interpolation}

Artificial intelligence has made significant progress in the application of ECG signal arrhythmia detection. Most works typically use synchronized 10-s complete ECG signal data, such as using CNN\cite{sau2025comparison,chen2024congenital,sangha2022automated}, CNN+RNN\cite{wang2021automated}, CNN+Transformer\cite{zhang202412} and hybrid models combining CNN and self-attention mechanism\cite{li2025electrocardiogram}, etc. {\color{blue}In recent years, Transformer-based methods have demonstrated outstanding performance in ECG classification. MSGformer\cite{ji2024msgformer} fuses multi-lead spatial features through self-attention and introduces multi-scale grid attention to capture temporal features at different scales. ECGTransForm\cite{el2024ecgtransform} combines multi-scale convolutions with a bidirectional Transformer to simultaneously capture spatial features and bidirectional temporal dependencies. AC-ITCT\cite{jeyarani2026interpretable} introduces a Temporal Convolutional Transformer with adaptive temporal resolution adjustment, balancing classification accuracy and interpretability. MDOT\cite{yisimitila2025bridging} converts one-dimensional ECG signals into two-dimensional oscillographic representations via an OSC module, incorporating clinical electrophysiological indicators and achieving lightweight deployment through momentum distillation. Chatterjee et al.\cite{chatterjee2026toward} utilized 8.2 million unlabelled ECGs for masked patch modeling pre-training, building a 1D Vision Transformer based on self-supervised learning. Yang et al.\cite{yang2025lead} systematically analyzed the impact of lead combinations on classification performance from a medical anatomical perspective, proposing a five-lead-grouping strategy and L5G-Net architecture.} When the input data dimensions are inconsistent, interpolation is usually used to uniformly fill the data to meet the input requirements of the model. Common interpolation strategies include zero padding, traditional machine learning interpolation methods, and deep learning based interpolation methods. SimMTM\cite{dong2023simmtm}, Medformer\cite{wang2024medformer}, S4D-ECG\cite{huang2024s4d}, CNN+SE\cite{li2025electrocardiogram} and methods such as BiLSTM\cite{liu2019mfb} use zero padding strategy to complete missing data before making predictions. These methods can effectively capture the connections between leads and model time-dependent features in synchronizing ECG signal data. Among them, Medformer\cite{wang2024medformer} introduces six data augmentation strategies to make it robust in the face of a certain degree of missing data. In addition, a Random Lead Masking (RLM) training strategy was proposed to ensure stable performance of the model even in the missing of complete leads\cite{liu2019mfb,chen2024congenital,zhu2026artificial}. By randomly masking a signal segment of about 1.5 seconds, the local loss caused by electrode detachment is simulated to improve the adaptability of the model to small-scale block like loss\cite{yao2020multi}. Using methods such as KNN and SAITS to interpolate missing data can to some extent complete data completion, allowing for subsequent prediction using existing synchronous 10s ECG signal analysis methods\cite{neog2025masking}.
However, the above methods all rely on interpolation and modeling strategies. Even with the use of the most advanced interpolation techniques, interpolation errors will still propagate in subsequent modeling, limiting the improvement of model performance. Meanwhile, the uncertainty introduced by interpolation can also reduce the interpretability of prediction results, posing potential challenges for clinical applications. {\color{blue}Moreover, the aforementioned Transformer methods all assume synchronized 10-second complete signals. When facing asynchronous 2.5s or 5s ECG data caused by different layouts, they cannot adapt to the temporal inconsistency among leads and large-scale Partial Blackout missing. For example, MSGformer relies on fixed-length time window partitioning, ECGTransForm requires complete temporal sequence context, MDOT's oscillographic conversion requires temporal alignment, and the PatchECG by Chatterjee et al.\cite{chatterjee2026toward}, although adopting a patch strategy, focuses on large-scale self-supervised pre-training without specifically modeling the Partial Blackout missing in asynchronous ECG.}

\subsubsection*{Modeling Directly Based on Missing Patterns}

MissTSM \cite{neog2025masking} proposed a new embedding scheme called Time Feature Independent (TFI) embedding, which considers each combination of time steps and features (or channels) as independent labels and encodes them into high-dimensional space, effectively capturing spatiotemporal dependencies without interpolation and achieving direct modeling of missing data. TimeXer introduces external information modeling to enable standard Transformers to coordinate the processing of relationships between internal and external sources \cite{wang2024timexer}. GRU-D proposed a GRU model with attenuation mechanism, which considers missing signals as external signals to enhance the modeling ability of the model on the original time series \cite{che2018recurrent}. BiTGraph further designed a biased temporal convolutional graph neural network to jointly model temporal dependencies and spatial structural features\cite{chen2023biased}. The above methods can directly model based on missing patterns and effectively capture potential temporal and multivariate dependencies.

However, these methods face serious challenges when dealing with asynchronous 2.5s and 5s ECG signals caused by different layouts. For example, RNN and GRU-D are prone to gradient vanishing when facing longer sequences; For a large number of continuous Partial Blackout missing in asynchronous ECG, TimeXer and MissSTM will generate a large number of continuous empty tokens, making it difficult for Transformer to effectively predict; In such scenarios, BiTGraph may also have a large number of continuous missing nodes, which affects the coherence and predictive performance of the graph structure. From this, it can be seen that in scenarios with different layouts of ECG, the above methods are difficult to accurately capture the collaborative dependencies and temporal features between leads, making it difficult to achieve effective modeling of ECG signals.


{\color{blue}
\subsubsection*{Multi-Layout ECG Processing Methods}
In recent years, research on multi-layout ECG processing has begun to emerge. Wang and Yang\cite{wang2025systematic} proposed a paper ECG digitization and classification framework combining adaptive rotated convolution with a state space augmented Transformer, capable of handling paper ECG images with rotations, creases, and high-level noise, but its classification model still predicts based on synchronized signals after digitization without modeling asynchronous missing. Yang et al.\cite{yang2025lead} analyzed the impact of lead combinations on classification performance from a medical anatomical perspective, proposing a five-lead-grouping strategy that reveals the importance of spatial complementarity among leads. These works provide important references for multi-layout ECG processing and understanding lead spatial relationships, but none address the signal asynchrony and large-scale Partial Blackout missing caused by different ECG layouts.
}

In summary, although progress has been made in ECG signal data imputation, direct modeling based on missing data, {\color{blue}and multi-layout ECG processing}, so far there has been no research attempting to solve the asynchronous signal and ``Partial Blackout'' missing problems caused by different layouts of ECGs. {\color{blue}Existing Transformer methods\cite{ji2024msgformer,el2024ecgtransform,jeyarani2026interpretable,yisimitila2025bridging} all assume temporally aligned and complete signals; the PatchECG by Chatterjee et al.\cite{chatterjee2026toward} focuses on large-scale self-supervised pre-training without designing specific mechanisms for asynchronous ECG; and the paper ECG digitization method\cite{wang2025systematic} only addresses the conversion from images to signals.} Therefore, this study proposes PatchECG, which designs a novel and universal architecture that adapts to signal data of different layouts, arbitrary lead numbers and lengths without introducing additional noise, and provides interpretability analysis of key patches. {\color{blue}Unlike existing methods, PatchECG directly models the Partial Blackout missing in asynchronous ECG through an adaptive variable block number missing learning mechanism and a patch-based collaborative dependency attention mechanism, effectively capturing spatiotemporal dependencies across leads without interpolation,} providing new insights for predicting cardiovascular diseases based on ECGs with different layouts.

\subsection*{Problem Definition}
Define multi lead ECG data as $X\in\mathbb{R}^{C \times T}$, where $C$ represents the number of ECG leads (electrodes) and $T$ represents the total timestamp. The input is ECG signal data, and the output is multiple possible disease or arrhythmia categories, which belong to the multi label and multi-classification task, that is, each input sample may correspond to multiple category labels at the same time.

\subsection*{Framework Overview}\label{overview}

\begin{figure*}[]
\centerline{\includegraphics[width=0.9\textwidth]{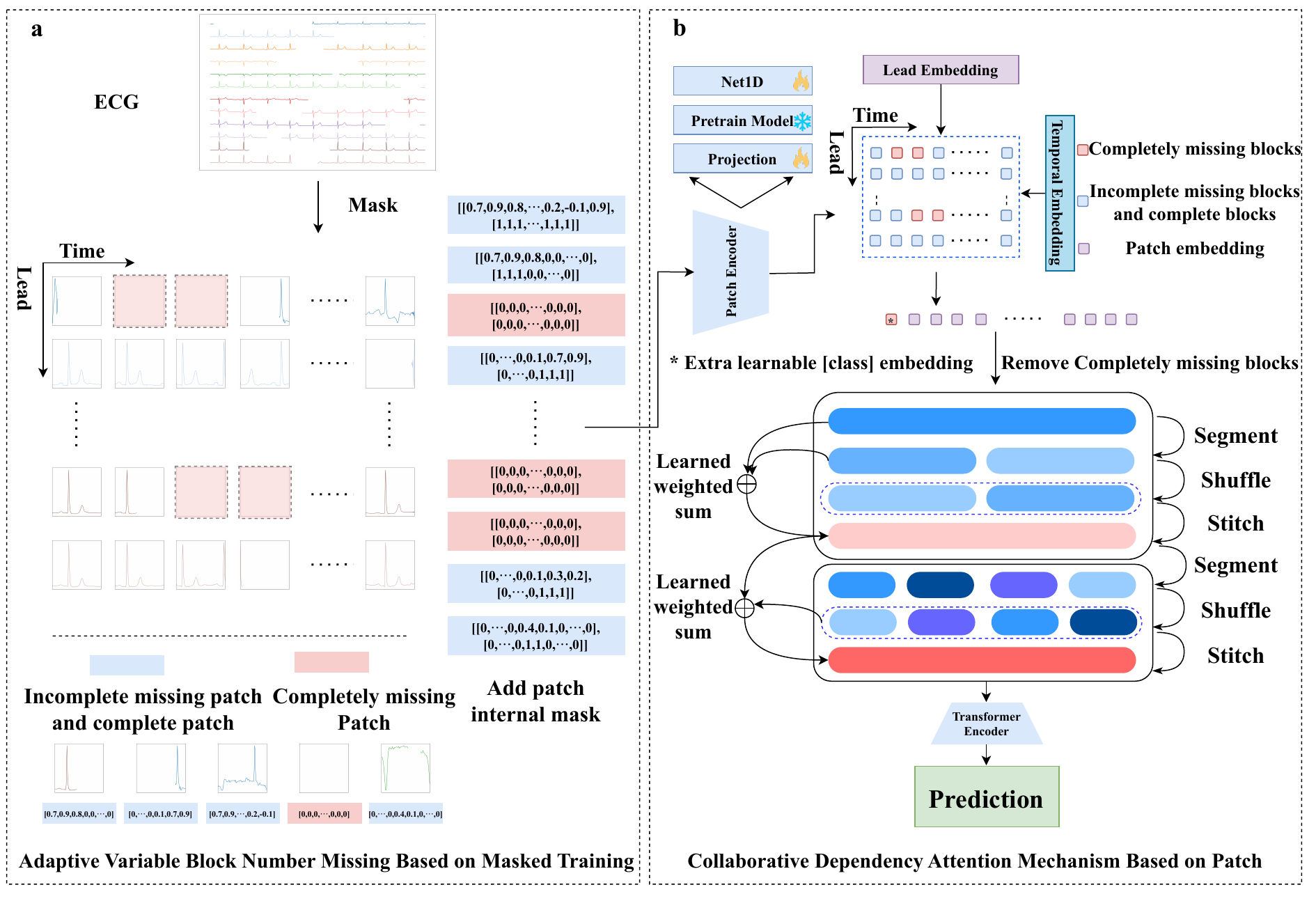}}
\caption{Framework Overview.}
\label{framwork-fig}
\end{figure*}

PatchECG is primarily composed of Patch Encoder(PatchEncoder), patch level temporal and lead embedding module, and patch oriented attention module. Figure \ref{framwork-fig} provides an overview of the PatchECG model proposed in this work.

\subsection*{Adaptive Variable Block Number Missing Based on Masked Training}

{\color{blue}
In order to enable the model to effectively learn different layouts of ECG, we propose a masking training strategy to improve the robustness of the model. Before training starts, each lead is independently masked by randomly sampling a starting position and a mask length. Specifically, for the $i$-th lead of total length $L$ (in sampling points), the mask starting position $st_i$ and mask length $l_i$ are independently drawn from uniform distributions:
\begin{equation}
  st_i \sim \mathcal{U}(0,\, L), \quad 
  l_i \sim \mathcal{U}(st_{i},\, L-st_{i})
\end{equation}
ensuring the masked region $[st_i,\, st_i + l_i)$ lies entirely within the signal boundary. Each lead contains at most one contiguous masked segment. All masked signal values are set to \texttt{NaN} to distinguish them from observed zero-valued signal points.

As shown in part a of the Figure~\ref{framwork-fig}, after masking, we perform equidistant non-overlapping segmentation on each lead using a fixed window size of $P$. For the $i$-th lead, the number of complete patches $n_i$ retained is:

\begin{equation}
  n_{i}=\left \lfloor \frac{c_{i}}{P}  \right \rfloor, \quad
  C\in \left \{  c_{i}\mid i=1,2\cdots C \right \}
\end{equation}
where $c_{i}$ denotes the total length of the $i$-th lead. The floor operation $\lfloor \cdot \rfloor$ ensures that only complete patches of exactly $P$ points are retained: when the remainder $r_i = c_i \bmod P > 0$, the last $r_i$ points are discarded without zero-padding or interpolation, so that no artificial values are introduced. The total patch count across all leads is $N = \sum_{i=1}^{C} n_i$.After segmentation, each patch $m$ is defined as:
\begin{equation}
m= \left \{  M_{i,j} \in \mathbb{R}^{P},\ 
i=1,\dots,C,\ j=1,\dots,\left \lfloor \frac{c_i}{P}  
\right \rfloor   \right \}
\end{equation}
Here $m$ denotes an individual patch. For each patch, we construct a binary mask indicator of the same length $P$, assigning a value of 1 to observed signal points and 0 to masked (missing) points. This indicator is then concatenated with the patch along the feature dimension to form the augmented patch $\tilde{m}$ is : 
\begin{equation}
\tilde{m} = \left \{  {\tilde{M}_{i,j}} \in \mathbb{R}^{2 \times P},\ 
i=1,\dots,C,\ j=1,\dots,\left \lfloor \frac{c_i}{P}  
\right \rfloor   \right \}
\end{equation}

Based on the sum of indicator values $\sum_k \mathbf{b}_{i,j}[k]$, each patch is classified into one of three types: a \textbf{complete patch} (all $P$ values observed), a \textbf{partially missing patch} ($0 < \sum_k \mathbf{b}_{i,j}[k] < P$), or a \textbf{completely missing patch} ($\sum_k \mathbf{b}_{i,j}[k] = 0$).
}

Input into PatchEncoder to obtain $\bar{m} \in \mathbb{R}^{D}$, where $D$ represents the embedding dimension output by PatchEncoder for each block. The existing signal-based methods include numerous excellent models that typically take fixed-layout and fixed-length signals as inputs. We effectively utilize these models by transforming them into PatchEncoders to encode inputs in patch form. In this way, PatchECG can process ECG signal data with any number of leads, any length, and any layout. We primarily employ the SE block with an attention mechanism as the basic module, combined with Net1D and a residual structure to extract high-dimensional features~\cite{hong2020holmes}.

\subsection*{Collaborative Dependency Attention Mechanism Based on Patch}
In our designed model, each block has two inherent features, different leads and different time positions. Unlike simple time series, we need to know the relative position of each block in the ECG. In order to enable PatchECG to better capture the relative position of each patch in the ECG, we initialized a lead embedding list $LE=\left \{ l_{0},l_{1},\dots,l_{C}  \right \} $ and a time embedding list $TE=\left \{ t_{0},t_{1},\dots,t_{\left \lfloor \frac{c_{i}}{P}  \right \rfloor }  \right \} $, Both can be learned during the training process. Add the lead code and time code to $\bar{m}$ to obtain: 

\begin{equation}
\bar{m}=\left \{ M_{i,j}+l_{i}+t_{j}\mid i=0,1,\dots C,j=0,1,\dots \left \lfloor \frac{c_{i}}{P}  \right \rfloor   \right \}
\end{equation}

{\color{blue}
To better capture the collaborative dependencies between non-adjacent patches across leads and time, we adopt the Segment, Shuffle, and Stitch (S3) module~\cite{grover2024segment} to rearrange the patch sequence $\bar{X} \in \mathbb{R}^{(N-N_{\text{miss}}) \times D}$ in a task-driven manner.

\textbf{Segment.} The input sequence $\bar{X}$ is divided into $n_s$ non-overlapping segments of equal length 
$\tau_s = \lfloor(N - N_{\text{miss}}) / n_s\rfloor$, yielding a segment set 
$\mathcal{S} = \{s_1, s_2, \dots, s_{n_s}\} \in \mathbb{R}^{\tau_s \times D \times n_s}$, where $s_k = \bar{X}[(k-1)\tau_s : k\tau_s]$. The initial number of segments is set to $n_s = 4$, following the ablation study in the original S3 work~\cite{grover2024segment}, which demonstrates that beginning with a coarse-grained partition 
and progressively refining it achieves a better trade-off between capturing global inter-lead dependencies and local temporal coherence. When $(N - N_{\text{miss}})$ is not exactly divisible by $n_s$, 
the first $(N - N_{\text{miss}}) \bmod n_s$ patches are truncated before segmentation and appended back to the output of the final S3 layer, ensuring no patch information is discarded.

\textbf{Shuffle.} A learnable shuffling vector $\mathbf{Q} = \{q_1, q_2, \dots, q_{n_s}\} \in \mathbb{R}^{n_s}$ 
determines the optimal reordering of segments. Each $q_k$ corresponds to segment $s_k$; segments are reordered according to the descending rank of $\mathbf{Q}$:
\begin{equation}
  \mathcal{S}^{\text{shuf}} = \text{Sort}(\mathcal{S},\ \text{key} = \mathbf{Q})
\end{equation}
To maintain differentiability, the permutation is implemented via a binary assignment matrix $\tilde{\Omega}$ derived from $\sigma_{\text{idx}} = \text{Argsort}(\mathbf{Q})$, allowing gradients to flow through $\mathbf{Q}$ during backpropagation~\cite{grover2024segment}. All elements of $\mathbf{Q}$ are initialized to the same constant value of $0.6$, so that the initial shuffling corresponds to the identity permutation; the optimal reordering is then learned jointly with the rest of the network in a task-driven manner.

\textbf{Stitch.} The shuffled segments are concatenated to reconstruct a full-length sequence 
$X^{\text{shuf}} \in \mathbb{R}^{(N-N_{\text{miss}}) \times D}$. To preserve the information in the original ordering while incorporating the newly learned arrangement, a learned weighted sum is performed via a Conv1D layer with learnable weights $w_1$ and $w_2$:
\begin{equation}
  \bar{X}^{\text{out}} = w_1 \cdot \bar{X} + w_2 \cdot X^{\text{shuf}}
\end{equation}
Both $w_1$ and $w_2$ are initialized to $1$ and updated jointly 
with all model parameters throughout training.

\textbf{Stacking.} The S3 module is stacked for $\varphi = 3$ layers to enable multi-granularity shuffling. In each subsequent layer $\ell$, the number of segments grows according to:
\begin{equation}
  n_s^{(\ell)} = n_s \times \theta^{\ell-1}, 
  \quad \ell = 1, 2, \dots, \varphi
\end{equation}
where $\theta = 2$ is the segment multiplier, yielding $n_s^{(1)} = 4$, $n_s^{(2)} = 8$, and $n_s^{(3)} = 16$ segments across the three layers. This progressive refinement allows the model to first capture coarse inter-lead dependencies at the block level, then refine to finer-grained intra-lead temporal structures. The output of the final S3 layer is $\bar{X}^{\prime} \in \mathbb{R}^{(N-N_{\text{miss}}) \times D}$, which encodes both the original and task-optimally reordered patch representations.
}

Before inputting into the transformer module, when dividing patches, we mark the blocks and record whether each block is completely missing, all of which are nan. If it is a completely missing block, we choose to discard it directly and only keep the complete and partially missing blocks. Obtain a variable length embedding sequence $\bar{X^{\prime}}$, similar to BERT. We add a learnable class embedding to obtain ${\bar{X}_{cls}}^{\prime}$:

\begin{equation}
{\bar{X}_{cls}}^{\prime}=[cls:\bar{X^{\prime}}]\in\mathbb{R}^{(N-N_{miss}+1)\times D}
\end{equation}

The Transformer Encoder contains many layers (denoted as K layers), and the output of the k-th layer is represented as:

\begin{equation}
\mathbf{H}^{(k)}=\text { TransformerEncoderLayer }\left(\mathbf{H}^{(k-1)}\right), \quad k=1,2, \ldots, K
\end{equation}

The initial input is:
\begin{equation}
\mathbf{H}^{(0)}={\bar{X}_{cls}}^{\prime}
\end{equation}

The final result is $\mathbf{H_{cls}}^{(K)}\in \mathbb{R}^{(N-N_{miss}+1)\times E}$ , where $E$ represents the output feature dimension. Finally, the prediction using a fully connected layer can be expressed as:

\begin{equation}
\hat{y} = \sigma  (\mathbf{H_{cls}}^{(K)}W+b),W \in \mathbb{R}^{E \times R},b\in \mathbb{R}^{R}
\end{equation}

Among them, $ W $ and $ b $ are trainable parameters, $ R $ represents the number of labels for multi label classification, $\sigma(\cdot)$ selects the sigmoid activation function to achieve probability output for multi label and multi-classification, and finally predicts the output:

\begin{equation}
\hat y \in \mathbb{R}^{R}
\end{equation}

\subsection*{Training}

All implementation details match the baseline. We use the exact hyperparameters of the baseline specified in the original paper, or when the code is available. Alternatively, when hyperparameters are not precisely specified in the paper or code, we attempt to maximize performance by searching for the best hyperparameters ourselves. During the training process of PatchECG, the window size $P$ is set to 64, the batch size is set to 64, and the number of training epochs is 30. The optimizer uses Adam, with an initial learning rate of $1 \times 10 ^ {-3}$ and a weight decay coefficient of $1 \times 10 ^ {-4}$. In the PatchEncoder module, the number of basic filters is set to 16, the size of the convolution kernel is 16, and the number of filters in each layer is [16, 32, 32, 40, 40, 64, 64]. Each layer contains two convolution blocks. The hidden feature dimension of the Transformer module is set to 768, with a depth of 3 layers and an 8-head attention mechanism. The S3\cite{grover2024segment} module consists of 3 layers, with an initial number of 4 segments and a segment multiplier of 2. Our code is implemented using PyTorch, and our experiments were conducted on NVIDIA RTX 3090Ti GPU.

\begin{table}[!h]
\centering
\caption{{\color{blue}Model Parameters and Inference Time Comparison.}}

\renewcommand{\arraystretch}{1.5}
\label{tab:model-complexity}
\begin{threeparttable}
\begin{tabular}{lcc}
\hline
\textbf{Model} & \textbf{Parameters (M)} & \textbf{Inference Time (ms)} \\
\hline
PatchECG      & 21.72  & 1.094               \\
ECG-Founder   & 30.82  & 23.900              \\
Medformer     & 12.11  & 1.350               \\
SimMTM        & \textbf{12.20}  & \textbf{0.021}               \\
SimMTM-KNN    & 12.20  & 4.872               \\
SimMTM-SAITS  & 13.55  & 1.319               \\
TimeXer       & 13.63  & 0.222               \\
\hline
\end{tabular}
\begin{tablenotes}
\small
\item Note: Bold values indicate the best performance in each column.
\end{tablenotes}
\end{threeparttable}
\end{table}

{\color{blue}
To assess the practical deployment feasibility of PatchECG, we report the model parameter count and per-sample inference time for all compared methods in Table~\ref{tab:model-complexity}. As shown, PatchECG contains 21.72M parameters and achieves an inference time of 1.094\,ms per sample, which is comparable to Medformer and SimMTM-SAITS, and substantially faster than ECGFounder. Although SimMTM has the shortest inference time (0.021\,ms), it achieves substantially lower AUROC across all layouts, indicating that its speed advantage comes at a significant cost to diagnostic accuracy. These results demonstrate that PatchECG strikes a favorable balance between model complexity, inference efficiency, and diagnostic performance, making it suitable for real-time clinical deployment scenarios.
}


\subsection*{Training Data}

\begin{figure}[!h]
\centerline{\includegraphics[width=0.8\textwidth]{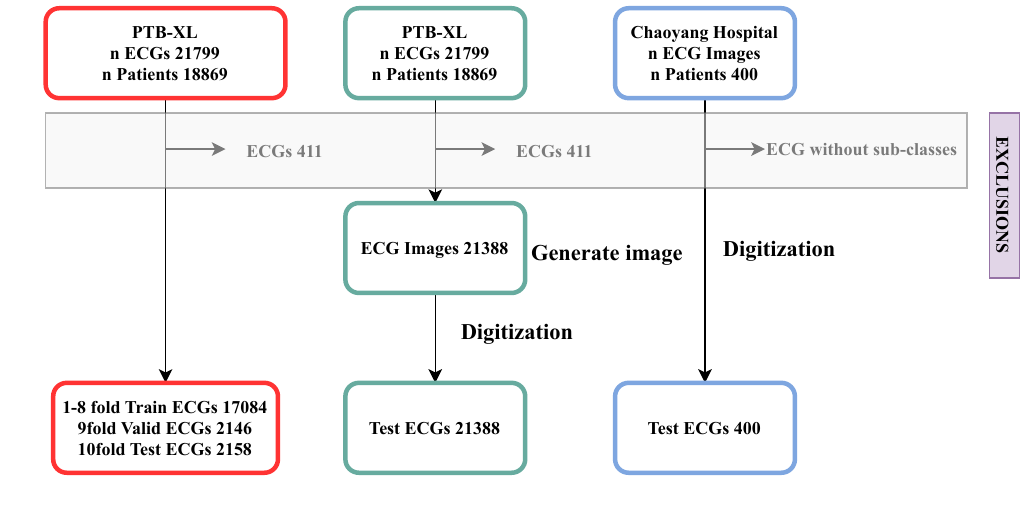}}
\caption{Data used in model development and validation for the ECG arrhythmia detection.}
\label{dataset}
\end{figure}

The data for this study comes from three sources, as shown in Figure \ref{dataset}.
\begin{itemize}
    \item The dataset used 21388 clinical 12-lead ECGs from 18869 patients using PTB-XL, with each sample lasting 10 seconds. 52\% are males and 48\% are females, with an age range of 0-95 years old \cite{PTB-XL}. We use multi label classification for this data: diagnostic subclasses. Specifically, the experiment used the official 10 fold split scheme provided by the PTB-XL dataset: in the first round, folds 1-8 were used as the training set, fold 9 as the validation set, and fold 10 as the test set; In the second round, the 7th fold is used as the validation set, the 8th fold is used as the test set, and the remaining folds are used as the training set; Repeat this process for the remaining rounds, rotating the division of the validation and testing sets to complete the evaluation of all folds. Use the Subclass (23 classes) of the multi-class classification task as the label.
    \item We used the tool provided by ECG-image-kit to generate 21388 ECG images with 3 x 4 layouts of different noise levels on the PTB-XL dataset, and further transformed these images into high-quality ECG signal data that can be used for subsequent analysis using the digital tool \cite{krones2024combining}, which ranked first in the 2024 CINC competition.
    \item To evaluate the generalization performance of the model, this study selected a cohort of 400 AF patients admitted to Chaoyang Hospital as an independent external validation set for performance verification\cite{tao2024artificial,tao2025multi}. {\color{blue} The clinical characteristics of this cohort are summarized in Table~\ref{tab:chatateristics}.} Due to differences in image resolution, size, and layout among the ECG data provided by Chaoyang Hospital, it is difficult to effectively process them using existing automated tools. Therefore, we used Paper ECG \cite{fortune2022digitizing} to manually digitize the image data to ensure the accuracy of the signal data.
    
\end{itemize}

\begin{table}[!h]
\centering
\caption{Clinical characteristics of the Chaoyang Hospital cohort.}
\renewcommand{\arraystretch}{1.5}
\label{tab:chatateristics}
\begin{threeparttable}

\begin{tabular}{lc}
\hline
\textbf{Characteristics} & \textbf{Chaoyang cohort} ($N=400$) \\
\hline
Age (years)                         & 67.1 $\pm$ 10.3          \\
Sex                                 &                          \\
\hspace{1em}{Male}                  & 189 (47.25\%)            \\
\hspace{1em}{Female}                & 211 (52.75\%)            \\
BMI (kg/m\textsuperscript{2})       & 26.2 $\pm$ 3.7           \\
Hypertension                        & 265 (66.25\%)            \\
Diabetes                            & 92 (23.00\%)             \\
Heart failure                       & 17 (4.25\%)              \\
Smoking status                      & 1 (0.25\%)               \\
Stroke                              & 1 (0.25\%)               \\
Atrial Fibrillation                 & 200 (50\%)              \\
Follow-up time (years)              & 4.82 $\pm$ 3.79          \\
\hline
\end{tabular}

\begin{tablenotes}
\small
\item Continuous variables (e.g., age and BMI) are presented as mean $\pm$ standard deviation. Categorical variables are presented as count (percentage).
\end{tablenotes}
\end{threeparttable}

\end{table}

\begin{figure}[h]
\centerline{\includegraphics[width=0.9\textwidth]{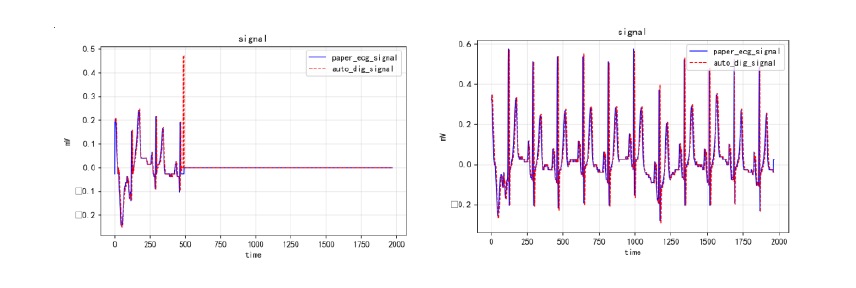}}
\caption{Overlay of original and digitized ECG signals.}
\label{snr-fig}
\end{figure}

{\color{blue}
Since the Chaoyang Hospital dataset contains only ECG image files without corresponding ground-truth digital signals, a direct evaluation of the Paper ECG digitization quality on this cohort is not feasible. To provide an indirect assessment of digitization accuracy, we conducted a controlled comparison using a separate subset of ECG images for which reference signals were available.

Specifically, we employed our in-house digitization algorithm, which incorporates a QRS waveform reconstruction strategy to address lead adhesion artifacts. Using this algorithm as a reference, we digitized \textbf{258} ECG images of fixed resolution covering both 3$\times$4 and 6$\times$2 layouts, and compared the resulting signals against those produced by the Paper ECG tool. Digitization quality was evaluated using two standard signal fidelity metrics:
}
\begin{equation}
\text{SNR} = 10 \log_{10} \frac{\sum_t x(t)^2}{\sum_t [x(t) - \hat{x}(t)]^2}
\end{equation}

\begin{equation}
\text{PSNR} = 10 \log_{10} \frac{x_{\max}^2}{\frac{1}{N}\sum_t [x(t) - \hat{x}(t)]^2}
\end{equation}
{\color{blue}
where $x(t)$ denotes the reference signal, $\hat{x}(t)$ the digitized signal, $x_{\max}$ the peak amplitude of the reference, and $N$ the total number of samples. Higher values of both SNR and PSNR indicate better digitization fidelity.

The results yielded an Avg SNR of $ \bm{-1.59}$ dB and an Avg PSNR of $ \bm{12.52} $ dB. We note that the relatively low numerical values do not necessarily reflect poor digitization quality, for two reasons. First, the reference signals used in this comparison are themselves derived from an independent digitization pipeline rather than ground-truth electronic recordings, introducing an inherent ceiling on achievable agreement. Second, minor temporal misalignments between the two digitized signals — even at the sub-sample level — can substantially penalize both SNR and PSNR due to their point-wise sensitivity to phase offsets. Visual inspection of the digitized waveforms confirms that the Paper ECG outputs closely follow the reference signals, with nearly identical morphology across all major waveform components, as illustrated in Figure~\ref{snr-fig}.
}

{\color{blue}
Figure~\ref{snr-fig} shows representative examples of the digitization quality comparison: panel \textbf{Left} presents a single \uppercase\expandafter{\romannumeral 1} Lead segment of 2.5 seconds, and panel \textbf{Right} presents a complete \uppercase\expandafter{\romannumeral 2} lead recording. In both panels, the blue trace represents the signal digitized by Paper ECG and the red trace represents the output of our in-house algorithm. The two signals exhibit strong visual agreement, supporting the conclusion that the Paper ECG tool produces reliable digitization output suitable for downstream model evaluation.
}

\subsection*{Evaluation Methods}

In our experiment, the task type is the multi-class classification. Specifically, during the model training and evaluation phase, we use the average Area Under the Receiver Operating Characteristic Curve (Avg AUROC) as the main evaluation metric; When conducting external validation on the AF dataset at Chaoyang Hospital, we comprehensively considered multiple indicators such as Accuracy, Precision, Recall, Specification, F1, and AUROC, with a particular focus on recall and specificity performance. In addition, due to significant label imbalance issues in the dataset used, we introduced the Focal Loss loss function \cite{lin2017focal} during the training process: $loss\left ( p \right )=-  \alpha _{t}{\left (  1-p\right ) }^{\gamma }\log_{}{p}$ To enhance attention to categories with fewer quantities and alleviate the adverse effects of category imbalance.

\section*{Results}
To comprehensively evaluate the performance of our proposed PatchECG, we conducted a series of rigorous experiments and compared it with several key baseline models, including: 1) The typical baseline \cite{neog2025masking} in the main follow MissTSM paper uses zero padding, KNN interpolation, and SAITS interpolation\cite{du2023saits}, respectively, and combines them with SimMTM for prediction \cite{dong2023simmtm}; 2) Medformer is a model specifically designed for biosignals, which introduces six different noise perturbations intra- and inter-granularity during the training phase, thus enhancing robustness to missing \cite{wang2024medformer};3) TimeXer, on the other hand, provides auxiliary cues to the model by explicitly modeling missing patterns and treating the missing as a kind of external information\cite{wang2024timexer}. We explored the diagnostic accuracy of the generated signal data digitized from ECG images by different layouts of signal data and the signal data digitized from different layouts of ECG images from the real Chaoyang Hospital.

\subsection*{PatchECG Achieve Better Results on Different Layouts}

\begin{figure*}[]
\centerline{\includegraphics[width=0.8\textwidth]{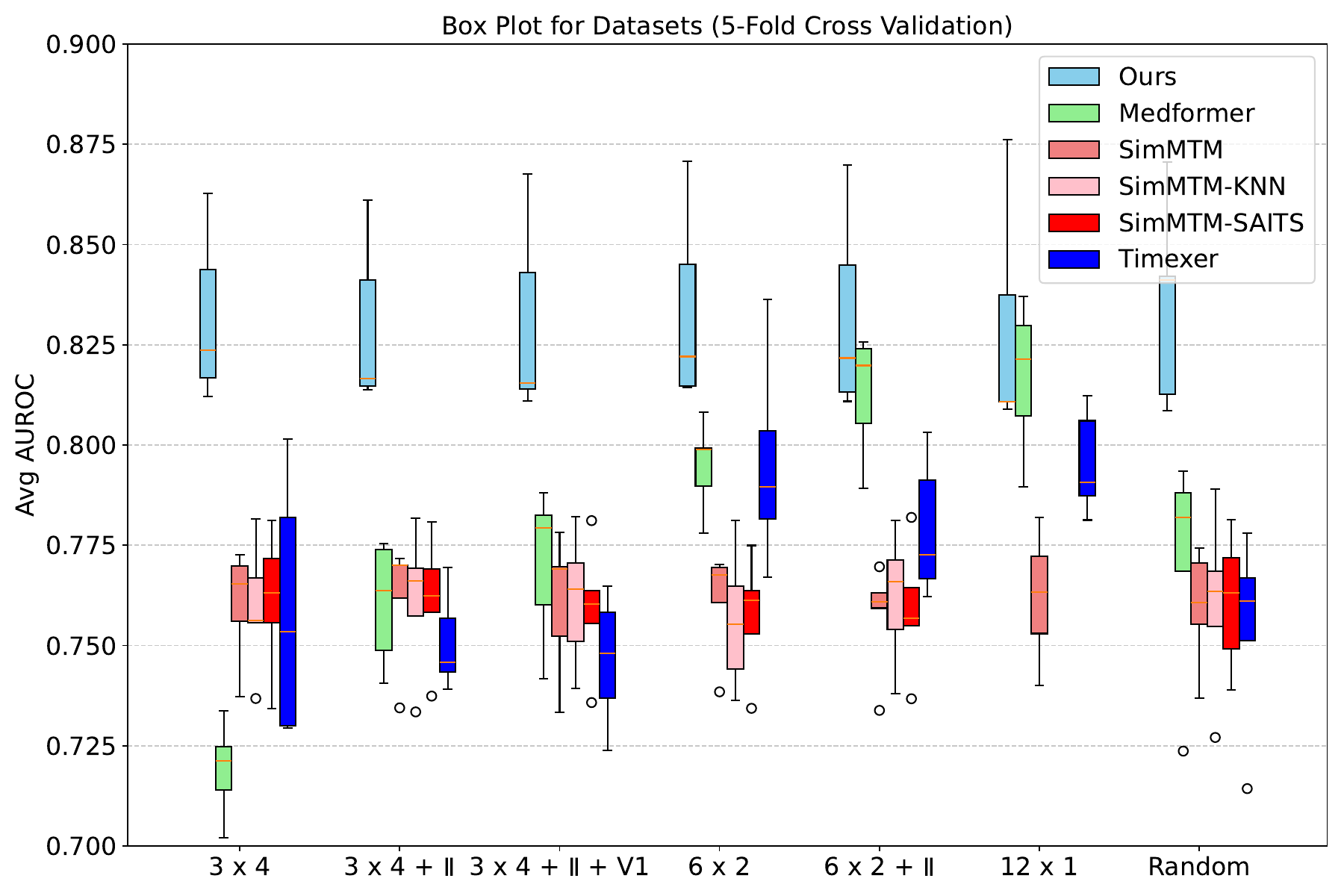}}
\caption{Results of PatchECG and Other Methods with Different Layouts. {\color{blue}Box plots represent the average AUROC performance across 5-fold cross-validation ($n=2158$ ECG samples per test fold). For each box plot, the center line indicates the median, the box limits indicate the upper and lower quartiles (interquartile range), and the whiskers extend to the minimum and maximum values.}}
\label{fig: figure_1}
\end{figure*}

As shown in Figure \ref{fig: figure_1}, our method obtained optimal average AUROC performance with 5-fold cross-validation under different ECG layouts. In the comparison with multiple baseline methods, we did not use the interpolation strategy due to the fact that interpolation may introduce hard-to-interpret noise and reduce the interpretability of the model. The experimental results show that the interpolation methods of zero padding, KNN and SAITS may introduce different degrees of noise interference under different layouts, and the model performance is similar, with the average AUROC mostly centered in the 0.750-0.775 range, and the different interpolation methods can not bring effective performance improvement. With the variation in lead layouts, Medformer performs better under more complete lead layouts, with its average AUROC showing an upward trend, and it performs comparably to our method in the 12 × 1 complete lead layouts.The performance of TimeXer is inferior in 3 × 4 layout compared to 6 × 2 layout. This is due to the missing of longer time segments in the 3 × 4 layout, leading to a higher number of consecutive token losses. Consequently, this affects the model's ability to capture temporal features, resulting in a relative decline in performance.

\subsection*{PatchECG is Highly Scalable}
{\color{blue}
To effectively utilize existing excellent models, we explored three different modules for encoding patches. We primarily adopted representative models such as Net1D, ECGFounder, and Projection as PatchEncoder:

\begin{itemize}
  \item \textbf{Projection}: Inspired by PatchTST~\cite{nie2022time}, it maps each patch to the embedding space through a single fully connected layer. This module has the simplest structure with no temporal feature extraction capability.
  
  \item \textbf{Net1D}: A residual one-dimensional convolutional model combined with an SE attention module~\cite{hong2020holmes}, trained from scratch on the target dataset. It is capable of learning local temporal features and channel-wise dependencies within each patch.
  
  \item \textbf{ECGFounder}: The latest large-scale ECG foundation model pre-trained on 10 million ECGs~\cite{li2025electrocardiogram}. Crucially, ECGFounder shares an identical backbone architecture with Net1D; the sole difference is that its parameters are frozen and directly transferred from large-scale pre-training, rather than trained from scratch on the target dataset.
  
\end{itemize}
}

\begin{table*}[htbp]
  \centering
  \caption{Different modules as PatchEncoder}
  \label{tab:patchEncoder}
  \begin{threeparttable}
  \sisetup{table-format=1.3}
  \scriptsize
  \begin{tabular*}{\textwidth}{
    l
    @{\extracolsep{\fill}}
    c c c c c c c
    c
  }
    \toprule
    \textbf{Method} & 
    \textbf{{\begin{tabular}{@{}c@{}}3 x 4 \end{tabular}}} &
    \textbf{{\begin{tabular}{@{}c@{}}3 x 4 + \uppercase\expandafter{\romannumeral 2} \end{tabular}}} &
    \textbf{{\begin{tabular}{@{}c@{}}3 x 4 + \uppercase\expandafter{\romannumeral 2} + V1 \end{tabular}}} &
    \textbf{{\begin{tabular}{@{}c@{}}6 x 2 \end{tabular}}} &
    \textbf{{\begin{tabular}{@{}c@{}}6 x 2 + \uppercase\expandafter{\romannumeral 2} \end{tabular}}} &
    \textbf{{\begin{tabular}{@{}c@{}}12 x 1 \end{tabular}}} &
    \textbf{{\begin{tabular}{@{}c@{}}Random \end{tabular}}} \\
    \midrule
    Ours-Projection & 0.768 $ \pm $ 0.034 & 0.781 $ \pm $ 0.037 & 0.782 $\pm$ 0.034 & 0.790 $\pm$ 0.032 & 0.792 $\pm$ 0.032 & 0.784 $\pm$ 0.028 & 0.780 $\pm$ 0.037 \\
    Ours-Net1D     & 0.831 $ \pm $ 0.019 & 0.829 $ \pm $ 0.019 & 0.830 $\pm$ 0.022 & 0.833 $\pm$ 0.022 & 0.832 $\pm$ 0.022 & 0.829 $\pm$ 0.026 & 0.835 $\pm$ 0.022 \\
    Ours-ECGFounder & \textbf{0.876} $ \pm $ \textbf{0.027} & \textbf{0.872} $ \pm $ \textbf{0.026} & \textbf{0.875} $\pm$ \textbf{0.028} & \textbf{0.879} $\pm$ \textbf{0.025} & \textbf{0.881} $\pm$ \textbf{0.026} & \textbf{0.882} $ \pm $ \textbf{0.024} & \textbf{0.872} $\pm$ \textbf{0.027} \\
    \bottomrule
  \end{tabular*}

\begin{tablenotes}
\small \centering
\item [] Note: Bold values indicate the best performance in each column.
\end{tablenotes}
\end{threeparttable}

\end{table*}

We adopted the same training, validation, and test set partitions as in Experiment 1 and conducted 5-fold cross-validation. As shown in Table~\ref{tab:patchEncoder}, all three PatchEncoder variants maintain consistent AUROC across different layouts, confirming that PatchECG effectively captures asynchronous signal dependencies regardless of layout configuration.
{\color{blue}
A detailed analysis of each module is as follows. \textbf{Projection} performs the worst due to its simple linear structure, which lacks the capacity to extract temporal features from patches. \textbf{Net1D} is trained from scratch and achieves the second-best result, demonstrating that a residual convolutional structure with SE attention can effectively learn patch-level representations. \textbf{ECGFounder} achieves the best average AUROC across all layouts. Notably, ECGFounder shares the same backbone architecture as Net1D; the sole difference is that its parameters are frozen weights transferred from pre-training on 10 million ECGs, rather than trained from scratch. This architectural equivalence enables a direct comparison: the consistent performance gain of approximately 0.04--0.05 AUROC over Net1D can be attributed specifically to the rich ECG morphology priors encoded by large-scale pre-training, rather than to any structural advantage. In addition, freezing the pre-trained parameters helps prevent overfitting, resulting in more stable performance across layouts.

In summary, PatchECG can effectively incorporate existing models as PatchEncoders, including large-scale pre-trained foundations. Combined with the adaptive variable block number missing mechanism, PatchECG maintains high accuracy across both asynchronous and synchronous ECG data with different layouts.
}

\subsection*{PatchECG Achieve Better Results on Generating ECG Images}

\begin{figure*}[]
\centerline{\includegraphics[width=0.8\textwidth]{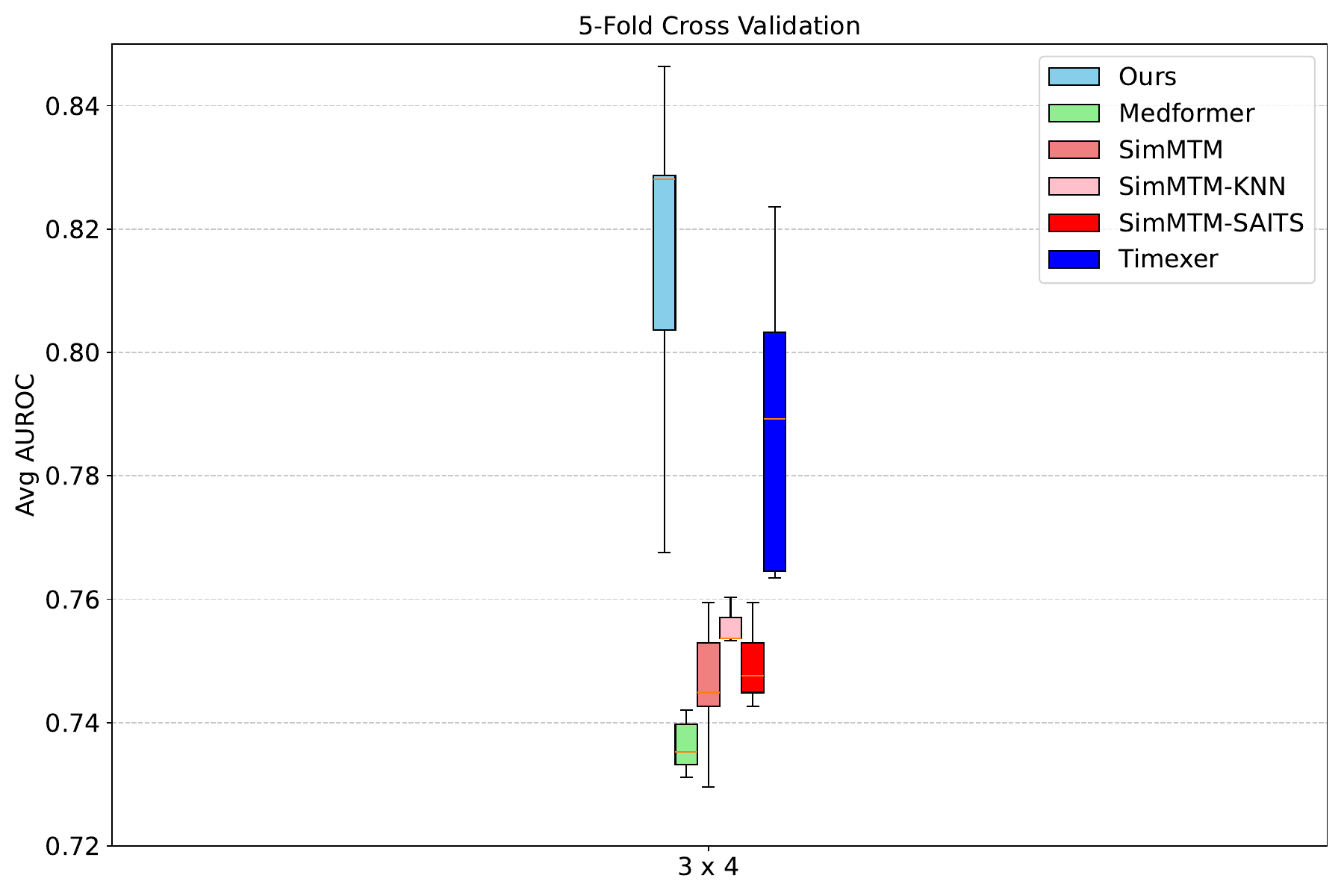}}
\caption{Digitally generated ECG image. {\color{blue}Box plots represent the average AUROC performance evaluated through 5-fold cross-validation ($n=21388$ digitally generated ECG samples). For each box plot, the center line indicates the median, the box limits indicate the upper and lower quartiles (interquartile range), and the whiskers extend to the minimum and maximum values.}}
\label{fig: figure_3}
\end{figure*}

To further verify the adaptability of the model in complex real-world scenarios, we generated ECG images using the ECG-image-kit tool. During the image generation process, common practical interference factors were simulated, including rotation, wrinkling, and light and shadow changes, to more realistically reproduce images encountered in clinical settings. {\color{blue} 
We adapted the ECG-image-kit tool to support the generation of ECG images across arbitrary layouts. However, the primary bottleneck in this experiment is not image generation but rather the absence of automated digitization algorithms capable of reliably handling arbitrary ECG layouts. We required a digitization solution that could provide both accurate signal reconstruction and reliable ground-truth labels at scale. The winning algorithm of the George B. Moody PhysioNet Challenge 2024~\cite{krones2024combining}, which represents the current state of the art in ECG digitization, best satisfies these requirements. Since the core focus of PatchECG is the downstream classification model rather than the digitization pipeline, we adopted this top-performing algorithm to ensure the highest input signal quality and avoid introducing additional digitization noise into the evaluation. This algorithm, however, only supports automated digitization of 3$\times$4 layout ECG images, which is why this experiment is restricted to the 3$\times$4 configuration.} Subsequently, the digitized signals were used as inputs for model testing, and the labels used in the testing phase were consistent with those used in the training phase. We adopted the AUROC as the evaluation metric and used 5-fold cross-validation to improve the stability and reliability of the results. The experimental results are shown in Figure~\ref{fig: figure_3}.

When the digitized signal obtained from the generated ECG image is used as the model input, our method performs the best among all baseline methods. Due to the high accuracy of digitization, TimeXer can effectively capture the exogenous information provided by the missing patterns, thus demonstrating excellent performance in this experiment. In contrast, MedFormer performs poorly under this layout, mainly because the 3 x 4 layout is the one that loses the most information, making it difficult for the model to obtain complete features, thus becoming the worst-performing baseline method in this experiment. Furthermore, other imputation-based baseline methods, whether using KNN imputation, SAITS imputation, or zero padding, perform almost equally well due to the introduction of additional noise in the missing areas. Different imputation methods do not bring significant improvements to the model.

\begin{figure}[h]
\centerline{\includegraphics[width=0.9\textwidth]{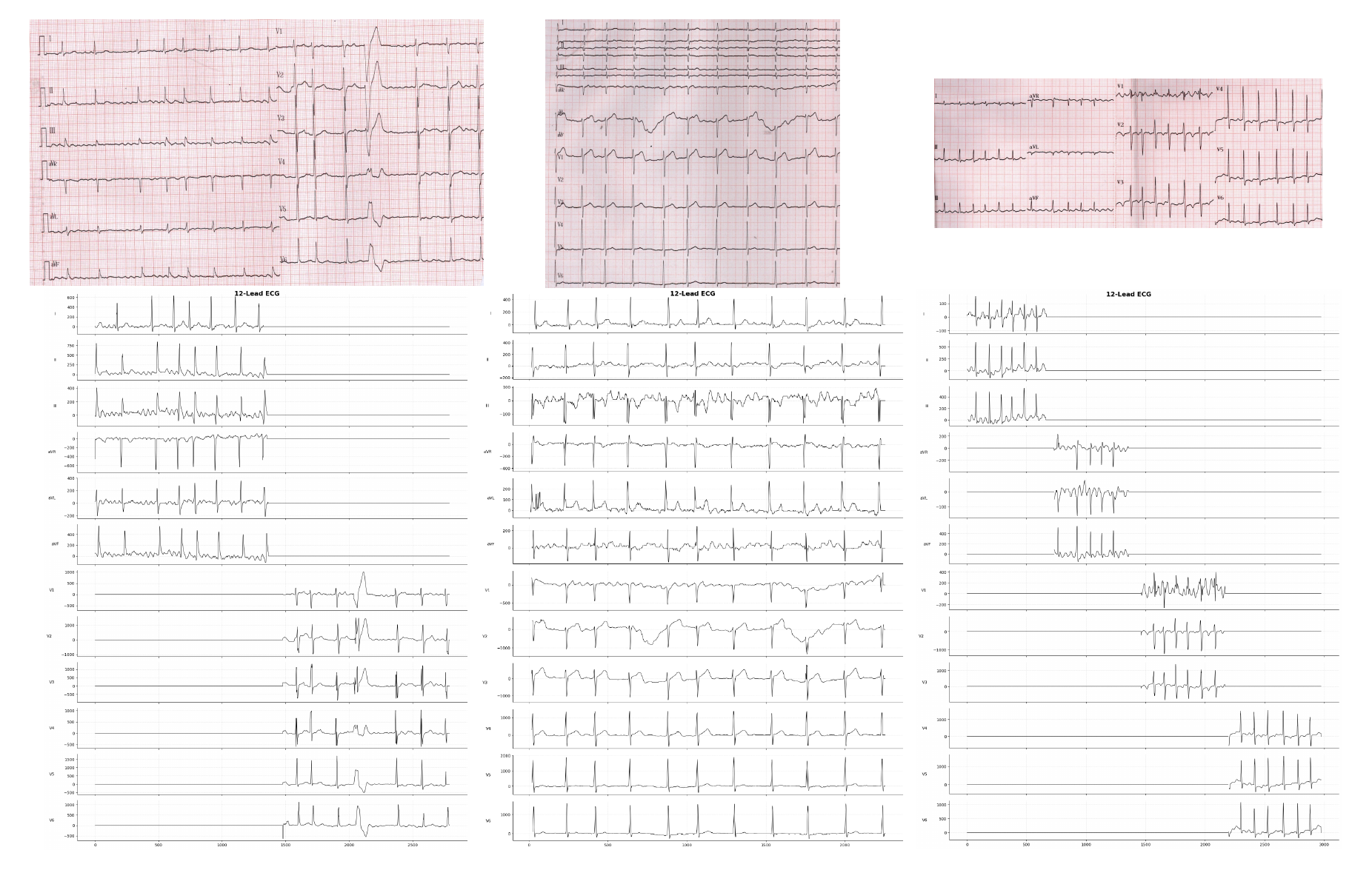}}
\caption{Examples of ECG images from the Chaoyang Hospital cohort and their corresponding digitized signals. From left to tight: 6$\times$2, 12$\times$1, and 3$\times$4 layouts.}
\label{dig-ecg-chaoyang}
\end{figure}

\subsection*{PatchECG Achieve Better Results on Real Hospital ECG Images}

On the external test set of Chaoyang Hospital, which mainly consists of 12 x 1 layout, with some being 3 x 4 and 6 x 2 layout (as shown in Figure \ref{dig-ecg-chaoyang}), our method achieved the best AUROC among all AF baseline predictions (as shown in Table \ref{tab:chaoyan_400}). The overall performance of ECGFounder's predictions is good. Although some methods have higher recall rates, their specificity is significantly lower, making them prone to misclassifying negative samples as positive, leading to an increased risk of misdiagnosis. Due to pre-training on tens of millions of ECGs, ECGFounder exhibits strong robustness. When there is a certain degree of data loss, the method fills in the data to a complete 12 x 1 layout through zero-padding before inputting it into the model. On the external AF cohort of Chaoyang Hospital, the three imputation strategies performed similarly in terms of AUROC scores, with almost no significant differences. This indicates that imputation does not significantly improve the model's accuracy. Medformer performed the best among all baseline methods. This method introduces six data augmentation strategies during the training stage, which act on its internal and external granularity modeling modules, respectively, to enhance the model's robustness to noise and missing data. When dealing with the Chaoyang Hospital cohort with different layouts and poor digitization quality, Medformer achieved an AUROC of 0.605, demonstrating strong adaptability and stability. TimeXer performed well on high-quality digitized signal data, but its performance significantly decreased in the Chaoyang Hospital cohort with poor digitization quality. This method treats missing data as exogenous information and uses it to assist in the modeling and prediction of endogenous information. However, when digitization is poor, not only is the endogenous information itself inaccurate, but the generated missing pattern information also becomes inaccurate. The exogenous information originally used to guide the model becomes a source of interference instead. The introduction of this noise weakens the model's judgment ability, leading to unsatisfactory prediction results on real Chaoyang Hospital image data.

\begin{table*}[htbp] 
  \centering
  \caption{Performance evaluation on an external test set at Chaoyang Hospital}
  \label{tab:chaoyan_400}
  \begin{threeparttable}
  \sisetup{table-format=1.4}
  \begin{tabular}{
    l
    S
    S
    S
    S
    S
    S
  }
    \toprule
    \textbf{Method} & {\textbf{Accuracy}} & {\textbf{Precision}} & {\textbf{Recall}} & {\textbf{Specificity}} & {\textbf{F1}} & {\textbf{AUROC}}\\
    \midrule
    Ours & \textbf{0.745} & \textbf{0.743} & \textbf{0.750} & \textbf{0.740} & \textbf{0.746} & \textbf{0.778}\\
    ECGFounder   & 0.625 & 0.605 & 0.720 & 0.530 & 0.658 & 0.667\\
    SimMTM       & 0.463 & 0.332 & 0.600 & 0.395 & 0.427 & 0.541\\
    SimMTM-KNN   & 0.485 & 0.480 & 0.540 & 0.430 & 0.512 & 0.539\\ 
    SimMTM-SAITS & 0.490 & 0.491 & 0.550 & 0.430 & 0.519 & 0.541\\
    Medformer    & 0.498 & 0.498 & 0.640 & 0.335 & 0.560 & 0.605\\
    TimeXer      & 0.489 & 0.487 & 0.460 & 0.515 & 0.473 & 0.485\\
    \bottomrule
  \end{tabular}

\begin{tablenotes}
\small \centering
\item [] Note: Bold values indicate the best performance in each column.
\end{tablenotes}
\end{threeparttable}
\end{table*}

\begin{table*}[htbp] 
  \centering
  \caption{Additional Evaluation of 12 x 1 layout on Chaoyang Hospital test set}
  \label{tab:chaoyan_300}
\begin{threeparttable}
  \sisetup{table-format=1.3}
  \begin{tabular}{
    l
    S
    S
    S
    S
    S
    S
  }
    \toprule
    \textbf{Method} & {\textbf{Accuracy}} & {\textbf{Precision}} & {\textbf{Recall}} & {\textbf{Specificity}} & {\textbf{F1}} & {\textbf{AUROC}}\\
    \midrule
    Ours & \textbf{0.833} & \textbf{0.723} & 0.810 & \textbf{0.845} & \textbf{0.764} & \textbf{0.893}\\
    ECGFounder & 0.573 & 0.430 & \textbf{0.860} & 0.430 & 0.573 & 0.703\\
    SimMTM     & 0.485 & 0.486 & 0.525 & 0.445 & 0.505 & 0.450\\
    SimMTM-KNN & \text{--} & \text{--} & \text{--} & \text{--} & \text{--} & \text{--} \\ 
    SimMTM-SAITS & \text{--} & \text{--} & \text{--} & \text{--} & \text{--} & \text{--} \\
    Medformer & 0.450 & 0.332 & 0.640 & 0.355 & 0.437 & 0.510\\
    TimeXer   & 0.473 & 0.289 & 0.390 & 0.510 & 0.331  & 0.423\\
    \bottomrule
  \end{tabular}

\begin{tablenotes}
\small \centering
\item [] Note: Bold values indicate the best performance in each column.
\end{tablenotes}
\end{threeparttable}

\end{table*}
Based on this, we further selected samples arranged in a 12 × 1 layout for additional evaluation (as shown in Table \ref{tab:chaoyan_300}). Our method still achieved the best performance in terms of the AUROC metric, demonstrating strong robustness.
{\color{blue}
ECGFounder achieved the highest recall (0.860) but the lowest specificity (0.430) in this evaluation. ECGFounder was applied directly to the Chaoyang Hospital cohort without fine-tuning or threshold recalibration; the threshold was adopted directly from the original paper. Having been pre-trained on over 10 million ECGs, the model may have developed a strong tendency toward AF recognition, causing it to over-detect AF-like patterns in non-AF samples when no target-domain adjustment is applied. The performance of Medformer improves with the completeness of lead data, showing notable improvement in the full 12$\times$1 layout with a AUROC of 0.51. In contrast, the other baseline methods largely fail under poor digitization quality, struggling to model or make accurate predictions. This further highlights the adaptability and advantages of PatchECG in real-world clinical 
environments.

It should be noted that none of the models in this study, including PatchECG, underwent threshold recalibration on the Chaoyang Hospital data, ensuring a fair comparison across all methods. Since recall and specificity are both threshold-dependent, the AUROC metric provides a more appropriate measure of intrinsic discriminative ability. ECGFounder's AUROC of 0.703 is substantially lower than PatchECG's 0.893, confirming that the gap reflects a genuine difference in model capability rather than a threshold artifact. 
}

\subsection*{Case Study}

\begin{figure*}[]
\centerline{\includegraphics[width=1\textwidth]{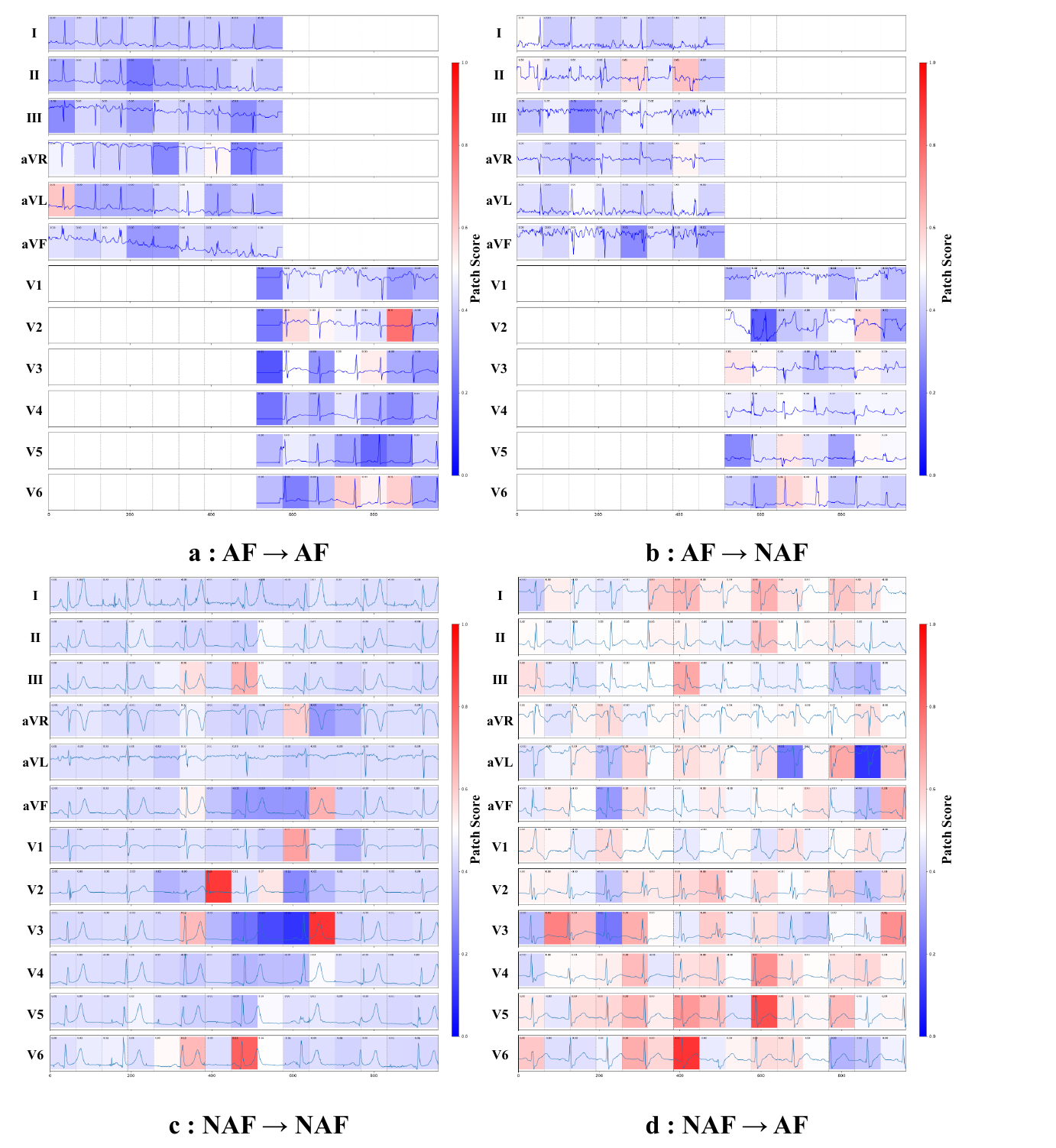} }
\caption{ \textbf{a}: Accurate prediction of atrial fibrillation; \textbf{b}: Misdiagnosis of atrial fibrillation as non-atrial fibrillation; this is a false negative; \textbf{c}: Successful identification of non-atrial fibrillation; \textbf{d}: Misdiagnosis of non-atrial fibrillation as atrial fibrillation; this is a false positive}
\label{fig:case_study}
\end{figure*}
As shown in Figure \ref{fig:case_study}, we selected ECG image data from Chaoyang Hospital and analyzed it based on its digitized signals. Specifically, it includes the following four typical cases:

(a) In images arranged in a 6 × 2 layout, the model is capable of accurately identifying key patches. For patches with high attention, a noticeable disappearance of P waves and the appearance of f waves can be observed, indicating that the model can accurately recognize AF characteristics in digitized asynchronous 5s ECG.

(b) Due to the poor quality of the digitalization method used, the digital signal data contains a large amount of noise, which severely interferes with the model mechanism. In this case, the patch that the model focuses on does not exhibit typical AF waveform characteristics, resulting in the misclassification of the AF sample as non-atrial fibrillation (NAF), which is a false negative.

(c) The model correctly classified a NAF sample as NAF. Upon further observation of the patches that the model focused on, we find that it pays attention to patches with atypical waveform changes in some leads, demonstrating its sensitivity to local abnormalities.

(d) In some images, due to poor digitization effects, there is a phenomenon of lead adhesion: the R wave of the lower lead may interfere with the S wave of the upper lead, causing severe jitter in the latter. At the same time, since the model is not trained on digitized data, it may misidentify this abnormal jitter as ST segment deviation. In addition, artifacts may also appear in the P wave region due to digitization distortion, and the model sometimes misidentifies this jitter as an f wave, thereby misdiagnosing NAF samples as AF.

\subsection*{Quantitative Evaluation of Model Interpretability}
{\color{blue}
To quantitatively assess whether the model's attention focuses on clinically relevant ECG segments, we conducted a comparison between the model's patch-level attention scores and annotations from two experienced cardiologists. Specifically, 20 ECG samples (10 AF and 10 NAF) were selected from the Chaoyang Hospital cohort. Each cardiologist independently identified the top 20 patches they considered most diagnostically relevant. The model's top 20 patches were determined by ranking the patch attention scores in descending order. Overlap was evaluated using two complementary metrics:

\textbf{Top-$K$ Overlap Rate (Top-$K$ OR)}, defined as the proportion of patches shared between the model's top-$K$ set and the clinician's top-$K$ set:
\begin{equation}
  \text{Top-}K\text{ OR} = \frac{|\mathcal{M}_K \cap 
  \mathcal{C}_K|}{K} \times 100\%
\end{equation}

\textbf{Jaccard Similarity Index (JSI)}, defined as:
\begin{equation}
  \text{JSI} = \frac{|\mathcal{M}_K \cap \mathcal{C}_K|}
  {|\mathcal{M}_K \cup \mathcal{C}_K|} \times 100\%
\end{equation}

where $\mathcal{M}_K$ and $\mathcal{C}_K$ denote the top-$K$ patch sets of the model and the clinician, respectively, and $K = 20$ in all experiments. Higher values indicate greater agreement.

The results are summarized in Table~\ref{tab:interp}. The inter-rater agreement between the two cardiologists serves as an upper-bound reference, yielding a Top-$K$ OR of 41.8\% $\pm$ 15.8\% and a JSI of 27.8\% $\pm$ 13.9\%, reflecting the inherent variability in clinical judgment even among experienced physicians. The model's agreement with Doctor~B reaches a Top-$K$ OR of 36.0\% $\pm$ 11.9\% and a JSI of 22.6\% $\pm$ 9.2\%, approaching the inter-rater ceiling. The lower agreement with Doctor~A (Top-$K$ OR: 20.5\% $\pm$ 10.7\%; JSI: 11.8\% $\pm$ 6.8\%) is consistent with the naturally high inter-clinician variability observed in patch-level ECG annotation tasks. These results collectively suggest that the model's attention mechanism captures diagnostically meaningful ECG segments that are broadly consistent with clinical expert judgment.

\begin{table}[h]
\centering
\caption{Quantitative comparison between model and 
cardiologist on ECG ($K = 20$).}
\label{tab:interp}
\begin{tabular}{lcc}
\toprule
\textbf{Comparison} & \textbf{Top-$K$ OR (\%)} & 
\textbf{JSI (\%)} \\
\midrule
Model vs Doctor A & $20.5 \pm 10.7$ & $11.8 \pm 6.8$ \\
Model vs Doctor B & $36.0 \pm 11.9$ & $22.6 \pm 9.2$ \\
Doctor A vs Doctor B & $41.8 \pm 15.8$ & $27.8 \pm 13.9$ \\
\bottomrule
\end{tabular}
\end{table}
}

\section*{Discussion}
The ECG is an essential tool in cardiological diagnosis, serving as a foundation for detecting and understanding various heart conditions. Despite the continuous advancements in digital ECG equipment, ECG images obtained through photographs, screen captures, or paper still prevail. With the continuous innovation of deep learning in medical research, several studies in the past two years have preliminarily explored methods for processing ECG data with different layouts. However, these methods often simply treat ECGs with different layouts as image classification problems and supplement them with image augmentation techniques. This approach may not only reduce prediction accuracy but also cast doubt on the credibility of model predictions among experts. Compared to image-based predictions, doctors and experts prefer arrhythmia detection based on ECG signals due to their more accurate interpretability. Thanks to advancements in digital methods, ECG image data can be accurately converted into ECG signal data, making it possible to effectively utilize these ECG image resources. Currently, there are numerous excellent research works based on ECG signal predictions, but when digitizing ECGs with different layouts into signals, specific challenges arise - existing methods do not fully consider how to solve the problems brought by signal data with different layouts. How to effectively bridge the gap between the original signal and the digitized signal and address the issue of different layouts will provide more comprehensive insights into ECG images and ECG signals for the future.

Based on the aforementioned concerns, we propose PatchECG, a novel predictive architecture that addresses signals with arbitrary layouts. It combines a masking training strategy and a patch-oriented attention mechanism to achieve intelligent recognition of arbitrarily layouts ECG arrhythmias. We systematically evaluated the proposed method on three types of data, including ECG data with different lead layouts, ECG image signals generated through digitization, and real clinical ECG images from Chaoyang Hospital. In all test scenarios, the model exhibited good predictive performance. Specifically, when training with raw signal data and testing on signal data after image digitization, the model still achieved stable and accurate prediction results\cite{sau2025comparison}. 
{\color{blue}The value of PatchECG lies in its ability to convert previously inaccessible ECG image archives into interpretable diagnostic outputs. By highlighting key signal patches with high attention scores, the model provides targeted decision support that aligns with how cardiologists examine ECGs—attending to specific waveform features across leads and time segments. This signal-grounded interpretability distinguishes our approach from image-based methods and is essential for clinical adoption.
}
In the external test set from Chaoyang Hospital, which had relatively poor quality, our method achieved an AUROC of 0.778 in the AF prediction task, demonstrating strong robustness. Furthermore, when evaluated on additionally filtered 12$\times$1 ECG data, the model's AUROC increased to 0.893. 
{\color{blue}
Notably, quantitative comparison with two experienced cardiologists (Table~4) shows that the model's patch-level attention achieves a Top-K overlap rate of up to 36.0\% and a Jaccard Similarity Index of 22.6\%, approaching the inter-clinician agreement ceiling (Top-K OR: 41.8\%; JSI: 27.8\%), which further validates the clinical relevance of the model's attention mechanism.
}
The interpretability of the model is of great significance for enhancing clinicians' understanding and trust in AI prediction results.

PatchECG provides a comprehensive and adaptable solution for clinical ECG image data and ECG signal data. This solution can accommodate signal data inputs with different layouts, arbitrary lead counts, and lengths. Its design philosophy is rooted in the diagnostic approach of doctors in ECG: when diagnosing specific diseases, doctors pay attention to abnormalities across leads and time segments simultaneously. Our model effectively focuses on key patches, identifying not only obvious abnormal segments in AF samples but also potential atypical waveform areas in some NAF samples. By visually focusing on key patches, the model can provide targeted decision support for clinicians, enhancing the interpretability and practical value of the model.

Regarding data usage and experimental setup, we use PTB-XL data as the training data source in this study. There are certain limitations in data usage, which to some extent restrict the model's generalization ability to broader patient populations and more complex pathological types. During the testing phase, we utilize the ECG-image-kit tool to generate ECG images and digitize them to evaluate the model. However, due to limitations in digitization methods, only 3 x 4 layout can be accurately processed, and other layout types of images cannot be effectively digitized. Therefore, in the generated images, we primarily focus on the 3 x 4 layout. In recent years, various digitization methods have emerged, but solutions that truly efficiently and accurately restore high-quality signals remain limited. It is particularly noteworthy that current digitization methods generally lack effective mechanisms to deal with lead adhesion. When lead adhesion occurs, the signals in the image overlap, which can easily be misjudged by the model as pathological features such as ST segment deviation, leading to deviations in prediction results. Therefore, how to improve the digitization quality of ECG images, especially developing digitization algorithms capable of identifying and correcting lead adhesion issues, is one of the important directions for achieving higher-quality AI-ECG diagnosis research in the future.

In the future, we will further integrate more high-quality and diverse ECG datasets to enhance the generalization ability and clinical applicability of the model. Additionally, given that PatchECG performs better on high-quality digital data, our team has developed the ability to partially reduce the impact of lead adhesion by reconstructing QRS waves. However, this technology is still difficult to fully apply to complex and variable ECG image scenarios. In the future, we will continue to advance the research and development of high-precision, low-noise ECG digitization technology, providing a more accurate signal foundation for downstream AI diagnosis.

\section*{Conclusion}


In this paper, we introduce PatchECG, a method that can be used to detect arrhythmias in ECG with different layouts. Specifically, we have designed an adaptive variable-block missing data representation learning mechanism, combined with a patch-oriented attention mechanism, which automatically focuses on key segments with collaborative dependencies in lead keys. Combined with spatiotemporal embedding, this enables the model to achieve accurate predictions in ECG data with different layouts without introducing additional noise. This method effectively fills the gap in current research regarding the diagnosis of ECGs with different layouts. At the same time, by enhancing the interpretability of key patches, it promotes collaboration between AI and clinicians, facilitating more efficient identification of arrhythmia.




\section*{Data availability}
The original PTB-XL dataset used in this study is publicly available on PhysioNet at \url{https://physionet.org/content/ptb-xl/1.0.3/}. The derivative dataset generated and analyzed during the current study—comprising 1-dimensional ECG signals reverse-digitized from synthesized multiple-layout ECG images (using the ecg-image-kit and the 2024 Computing in Cardiology Challenge 1st place digitization algorithm)—is publicly available in the Zenodo repository at \url{https://zenodo.org/records/19254077}. The diagnostic labels for these digitized signals correspond directly to the ground-truth annotations of the original PTB-XL dataset.The Beijing Chaoyang Hospital Institutional Research Board approved the study protocol and all ethical considerations (No. 2021-Ke-555). The need for informed consent was waived by the ethics board of our hospital because the images had been acquired during daily clinical practice. 


\section*{Code availability}
The imolementation of PatchECG is available in the GitHub repositorv at \url{https://github.com/zhangshanwei-mm/PatchECG}. PaperECG is available in the GitHub repository at \url{https://github.com/Tereshchenkolab/paper-ecg/tree/master}. Our internal digitization algorithms are available in the GitHub repository at \url{https://github.com/PKUDigitalHealth/ecg-img2ts}.The open-source ecg-image-kit tool utilized for generating ECG images is available a \url{https://github.com/alphanumericslab/ecg-image-kit}. The automated digitization of ECG image into digital signals was performed using ECG Digitiser, the 1st-place automated digitization algorithm in the George B. Moody PhysioNet Challenge 2024. The code is available at \url{https://github.com/felixkrones/ECG-Digitiser}.

\bibliography{reference}

\section*{Acknowledgements} 


This work was supported by the National Natural Science Foundation of China (Grant 62102008 to SH; Grant 62376197, 62020106004 and 92048301 to YZ); CCF - Tencent Rhino-Bird Open Research Fund (Grant CCF-Tencent RAGR20250108 to SH); CCF-Zhipu Large Model Innovation Fund (Grant CCF-Zhipu202414 to SH); PKU-OPPO Fund (Grant BO202301 and BO202503 to SH);Research Project of Peking University in the State Key Laboratory of Vascular Homeostasis and Remodeling (Grant 2025-SKLVHR-YCTS-02 to SH); Beijing Municipal Science and Technology Commission (Grant Z251100000725008 to SH); Tianchi Elite Youth Doctoral Program (Grant CZ002701 and CZ002707 to YZ).

\section*{Author information}
\subsection*{Authors and Affiliations}
\textbf{Department of Computer Science, Tianjin University of Technology, Tianjin, China}

\noindent Shanwei Zhang, Kexin Wang, Yuxi Zhou \& Shengyong Chen

\noindent \textbf{HeartVoice Medical Technology, Hefei, China}

\noindent Deyun Zhang \& Shijia Geng

\noindent \textbf{Department of Cardiology, Beijing Chaoyang Hospital, Capital Medical University, Beijing, China}

\noindent Yirao Tao \&  Xingpeng Liu

\noindent \textbf{Department of Cardiology, Peking University People’s Hospital, Beijing, China}

\noindent Qinghao Zhao

\noindent \textbf{DCST, BNRist, RIIT, Institute of Internet Industry, Tsinghua University, Beijing, China}

\noindent Yuxi Zhou

\noindent \textbf{Tianjin Key Laboratory of Ionic-Molecular Function of Cardiovascular Disease, Department of Cardiology, Tianjin Institute of Cardiology, the Second Hospital of Tianjin Medical University, Tianjin 300211, China}

\noindent  Xingliang Wu

\noindent \textbf{National Institute of Health Data Science, Peking University, Beijing, China}

\noindent Shanwei Zhang, Jun Li \& Shenda Hong

\noindent\textbf{Institute for Artificial Intelligence, Peking University, Beijing, Chin}

\noindent Shenda Hong

\noindent \textbf{Institute of Medical Technology, Peking University Health Science Center, Beijing, China}

\noindent Shenda Hong

\noindent \textbf{State Key Laboratory of Vascular Homeostasis and Remodeling, NHC Key Laboratory of Cardiovascular Molecular Biology and Regulatory Peptides, Peking University, Beijing, China}

\noindent Shenda Hong

\subsection*{Corresponding author}
Correspondence to Shenda Hong \& Yuxi Zhou.

\section*{Contributions}
S.Z. performed the primary data processing, algorithm implementation, and drafted the initial version of the manuscript.Q.Z. and X.W. provided clinical expertise and conducted the medical evaluation of the model's interpretability. D.Z., Y.T., and K.W. contributed to data aggregation, baseline preparation. S.G., J.L., and X.L. assisted with the design of the validation framework and carried out additional data analyses.  S.C. contributed to the conceptualization of the methodology and co-developed the proposed solution through critical discussions. Y.Z. and S.H. conceived and supervised the overall study, provided critical revisions to the manuscript, and secured access to data and computational resources. All authors critically reviewed and approved the final manuscript.

\section*{Ethics declarations}
\subsection*{Competing interests}
The authors declare no competing interests.

\bigskip


%

\begin{appendices}

\section{Manual Digitization Procedure Using Paper ECG}\label{appendixA}
{\color{blue}
The Chaoyang Hospital ECG images were digitized using the Paper ECG tool~\cite{fortune2022digitizing}(available at \url{https://github.com/Tereshchenkolab/paper-ecg/tree/master}), an open-source software capable of digitizing ECG images across arbitrary layouts through a semi-automated, operator-guided procedure. The detailed digitization workflow is illustrated in Figure~\ref{paperecg_workflow}, and consists of the following steps.

\begin{figure}[h]
\centerline{\includegraphics[width=0.9\textwidth]{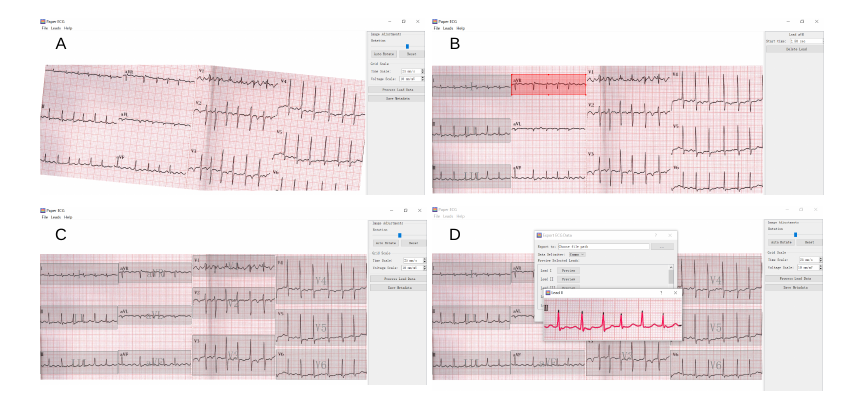}}
\caption{\textbf{PaperECG digitization.} (A) Rotate ECG image and configure paper speed and voltage gain. (B) Set the baseline starting position for each lead. (C) Select all leads for digitization. (D) Preview and execute the digitization process.}
\label{paperecg_workflow}
\end{figure}

\textbf{Step 1 — Image preprocessing.} The ECG image is first rotated to correct for any scanning skew. The operator then manually specifies the paper speed (typically 25 mm/s) and voltage calibration (typically 10 mm/mV) according to the scale markings visible on the ECG image.

\textbf{Step 2 — Lead boundary annotation.} For each lead, the operator manually places the lead border markers at the corresponding waveform region in the image. The lead onset time is set according to the ECG layout:

\begin{itemize}
  \item \textbf{3$\times$4 layout}: Leads I, II, and III start at 0\,s; leads aVR, aVL, and aVF start at 2.5\,s; leads V1--V3 start at 5.0\,s; leads V4--V6 start at 7.5\,s.
  
  \item \textbf{6$\times$2 layout}: The first six leads start at 0\,s; the remaining six leads start at 5.0\,s.
  
  \item \textbf{Rhythm strip (Lead II)}: When a full-length rhythm strip is present, it is annotated separately with an onset time of 0\,s.
\end{itemize}

\textbf{Step 3 — Signal extraction.} Once all lead boundaries and onset times are configured, the tool automatically extracts the digitized one-dimensional signal for each lead based on the operator-defined annotations.

The semi-automated nature of this procedure requires consistent operator practice to ensure reproducibility. All digitization in this study was performed by trained operators following the standardized protocol described above.

}









\end{appendices}
\end{document}